# CHARACTERIZATION AND MITIGATION OF INSUFFICIENCIES IN AUTOMATED DRIVING SYSTEMS

**Yuting Fu**[1], **Jochen Seemann**[1], **Caspar Hanselaar**[2], **Tim Beurskens**[1], **Andrei Terechko**[1], **Emilia Silvas**[2,3], **Maurice Heemels**[2]

[1] – Central Technology Office, NXP Semiconductors, Eindhoven, The Netherlands
[2] – Department of Mechanical Engineering, Eindhoven University of Technology, Eindhoven, The Netherlands
[3] – Integrated Vehicle Safety, TNO, Dutch Organization for Applied Scientific Research, Helmond, The Netherlands

# ABSTRACT

Automated Driving (AD) systems have the potential to increase safety, comfort and energy efficiency. Recently, major automotive companies have started testing and validating AD systems (ADS) on public roads. Nevertheless, the commercial deployment and wide adoption of ADS have been moderate, partially due to system functional insufficiencies (FI) that undermine passenger safety and lead to hazardous situations on the road.

In contrast to system faults that are analyzed by the automotive functional safety standard ISO 26262, FIs are defined in ISO 21448 Safety Of The Intended Functionality (SOTIF). FIs are insufficiencies in sensors, actuators and algorithm implementations, including neural networks and probabilistic calculations. Examples of FIs in ADS include inaccurate ego-vehicle localization on the road, incorrect prediction of a cyclist maneuver, unreliable detection of a pedestrian in rainy weather using cameras and image processing algorithms, etc.

The main goal of our study is to formulate a generic architectural design pattern, which is compatible with existing methods and ADS, to improve FI mitigation and enable faster commercial deployment of ADS. First, we studied the 2021 autonomous vehicles disengagement reports published by the California Department of Motor Vehicles (DMV). The data clearly show that disengagements are five times more often caused by FIs rather than by system faults. We then made a comprehensive list of insufficiencies and their characteristics by analyzing over 10 hours of publicly available road test videos. In particular, we identified insufficiency types in four major categories: world model, motion plan, traffic rule, and operational design domain. The insufficiency characterization helps making the SOTIF analyses of triggering conditions more systematic and comprehensive.

To handle faults, modern ADS already integrate multiple AD channels, where each channel is composed of sensors and processors running AD software. Our characterization study triggered a hypothesis that these heterogeneous channels can also complement each other's capabilities to mitigate insufficiencies in vehicle operation. To verify the hypothesis, we built an open-loop automated driving simulation environment based on the LG SVL simulator. Three realistic AD channels (Baidu Apollo, Autoware.Auto, and comma.ai openpilot) were tested in the same driving scenario. Our experiments suggest that even advanced AD channels have insufficiencies that can be mitigated by switching control to another (possibly less advanced) AD channel at the right moment.

Based on our FI characterization, simulation experiments and literature survey, we define a novel generic architectural design pattern Daruma to dynamically select the channel that is least likely to have a FI at the moment. The key component of the pattern does cross-channel analysis, in which planned trajectories and world models from different AD channels are mutually evaluated. The output of the cross-channel analysis is combined with more traditional fault detections in a safety fusion component. The safety fusion then feeds an aggregated per-channel safety score to the high-level arbiter, which eventually selects the AD channel to control the vehicle. The formulated architectural pattern can help manufactures of autonomous vehicles in mitigating FIs.

Limitations of our study suggest interesting future work, including algorithmic research on cross-channel analysis and safety fusion, as well as evaluation of the cross-channel analysis in simulations and road tests.

# 1. INTRODUCTION

Potential advantages of Automated Driving (AD) systems include increased safety, comfort and energy efficiency [1]. Major automotive companies and academic institutions have already started testing and validating AD systems (ADS) on public roads [2, 3, 4, 5]. Nevertheless, the commercial deployment and wide adoption of ADS have been limited, partially due to the challenge of functional insufficiencies (FIs), which is the focus of this paper.

Traditionally, safety-critical systems focused on detecting and mitigating faults, such as a memory cell bit flip or a software deadlock, which are the primary focus of the automotive standard ISO 26262 [6]. Modern ADS deploy redundant AD channels to cope with faults [7]. As illustrated in Figure 1, an AD channel consists of the sensors (such as radar and camera), the ADS display for visualization and human interaction, and the processing modules including perception, localization, maps, prediction, and motion planning module, but excludes actuators for steering and acceleration. The greyed out items in Figure 1 are the human machine interface and the actuators, which are shared among the multiple AD channels of the ADS.

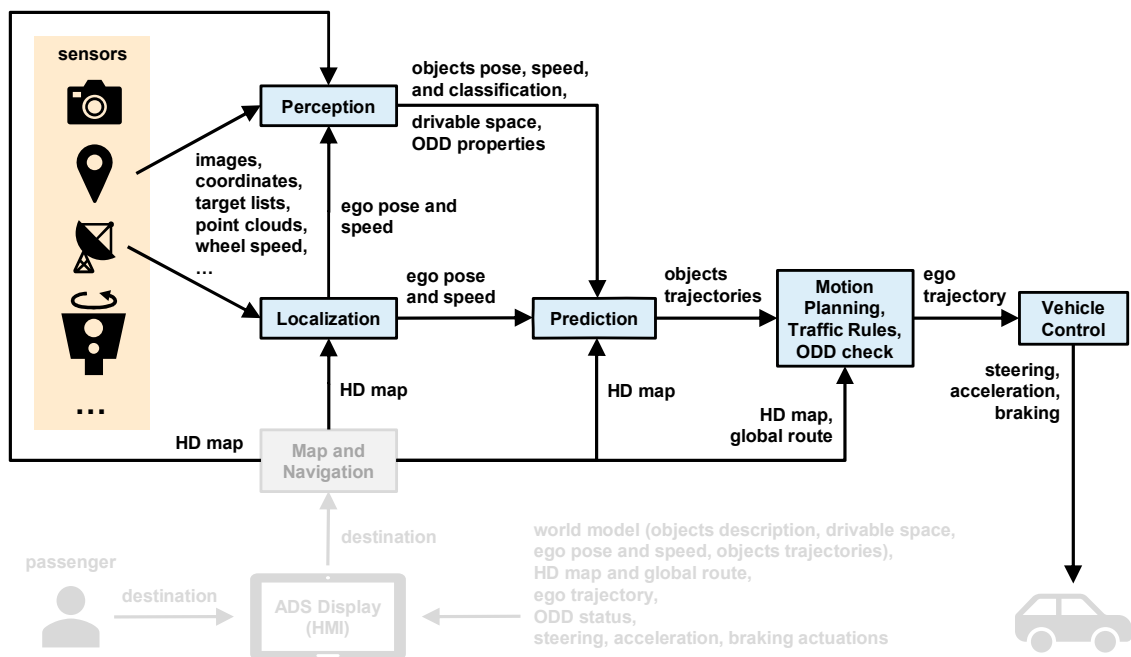

**Figure 1. The simplified AD channel architecture.**

In our work we do not focus on fault handling, because it has been widely studied and handled by effective safety mechanisms leveraging error-correction codes, watchdog monitors, redundancy, etc. In contrast, advanced environmental sensors, deep neural networks, and probabilistic or non-deterministic algorithms in modern ADS introduce insufficiencies in the implementation of the AD functionalities which have received significantly less attention. Such insufficiencies are called FIs in the ISO 21448 standard Safety Of The Intended Functionality (SOTIF) [8].

An example of a FI is illustrated in Figure 2 below, where the ADS is turning right on the T-shaped intersection and does not properly predict the intention of the pedestrian walking across the street. Consequently, the ADS does not slow down [9]. To avoid collision the human driver overrides the ADS operation and takes over control of the vehicle. This example show-cases the potential hazardous consequences of FIs, as well as the reduced ADS availability due to the disengagement. Note that the ADS and the ego vehicle in this example do not suffer from a fault in the traditional sense of functional safety as defined in the standard ISO 26262 [6].

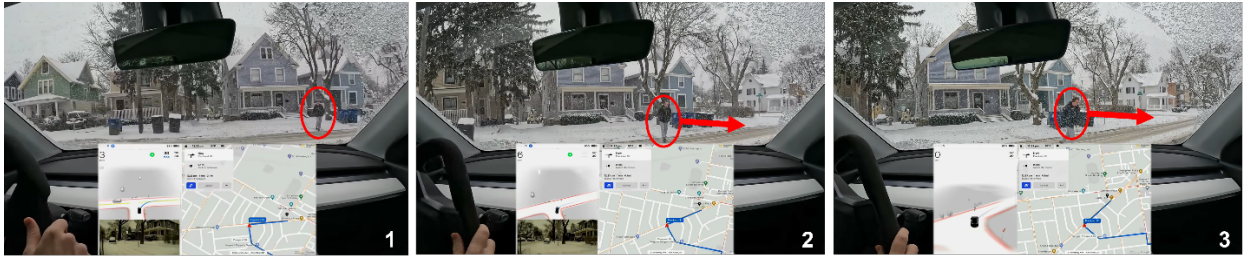

**Figure 2. Incorrect prediction of a pedestrian's trajectory is an example of ADS functional insufficiencies. The camera-based perception module of the ADS correctly detects and tracks the pedestrian in step 1 and step 2. However, in step 2 it fails to predict the pedestrian's intention to cross the street, and in step 3 the human driver disengages automated driving.**

Some other insufficiency examples are:

1. ADS perception fails to detect a pedestrian;
2. ADS prediction fails to correctly predict the cyclist's maneuver;
3. ADS motion planning fails to yield a trajectory feasible for vehicle actuators;
4. ADS drives the vehicle faster than the speed limit.

Mitigating FIs at runtime with the help of design patterns is the primary focus of this paper. Other approaches, such as product lifecycle management and various other safety techniques recommended by SOTIF at design time are outside of the scope of this paper. A classic example of a FI mitigation technique is sensor fusion [10]. It integrates multi-modal sensors output into a single world model by combining complementary sensor data with varying weights dependent on the situation. For example, the camera sensor will be favored in a clear weather condition, while the radar sensor will be preferred in rainy conditions. Note that these sensors can also have different capabilities, for example, the camera can perceive the traffic light color, while the radar sensor can detect object speeds.

Our work concentrates on two essential problems and contributes to their solutions:

1. There is no comprehensive inventory and characterization of FIs in modern ADS. Providing such an inventory will be beneficial in assessing the prevalence of FIs and their potential impact on ADS safety. Our paper presents a new comprehensive characterization of FIs in ADS, which can be used in the analysis of triggering conditions recommended by the SOTIF.
2. Despite integration of state-of-the-art functional safety mechanisms and redundant AD channels for fail-operational behavior, modern ADS still occasionally suffer from FIs that compromises safety and availability. Consequently, how to reduce the negative impact of FIs in autonomous safety-critical systems [11, 12] is an important problem. We contribute to its solution by introducing Daruma: a generic architectural design pattern extending redundant ADS with cross-channel analysis and arbitration of redundant AD channels to mitigate FIs.

It should be noted that the high-level goal of our study is to support technological evolution to safe AD and reduce road accidents. We do not intend to assess safety of the evaluated ADS, but rather to extract architectural recommendations to reduce the negative impact of FIs.

The remainder of this paper is organized as follows. In the next section we survey literature on FIs and architectural design patterns to mitigate them. Section 3 discusses analysis of our road test studies, presents our characterization table of FIs, and makes a hypothesis about a new architectural design pattern to mitigate FIs. The open-loop simulation method and results in Section 4 are used in the subsequent section to define the design pattern for cross-channel analysis. In Section 5, we conclude our findings and discuss future research directions in Section 6.

## 2. RELATED WORK

In this section we first present a summary of related work on reports and statistical analyses of FIs, then introduce several existing FI mitigation methods.

### ANALYSIS OF FUNCTIONAL AND OUTPUT INSUFFICIENCIES

As stated in [13], hazardous situations caused by FIs are typically not considered by the traditional safety analysis process focusing on ISO 26262 [6]. To identify and evaluate such hazards, SOTIF [8] focuses on identifying triggering conditions of the FIs and the acceptance criteria of the ADS response in the presence of the triggering conditions. Triggering conditions as defined in SOTIF, such as unusual road users, weather conditions, light glare, EMI interference, etc., are external factors that initiate the ADS FIs. Furthermore, internal ADS implementation insufficiencies can also cause FIs [8, 14, 15, 16]. Representation of the external scene and ego vehicle behavior are captured in the state of the ADS components. Erroneous component state, such as missed object or ghost object detection, incorrect predictions of trajectories, etc., is termed output insufficiencies (OI) in SOTIF. Our characterization study focuses on OIs to make a comprehensive characterization of FIs in ADS. The characterization can help employ appropriate mitigation mechanisms to address a wider range of triggering conditions. Note that ISO 26262, SOTIF and UL 4600 [17] standards all focus on technology-neutral design-time processes, while our work concentrates on concrete FIs and a safety mechanism to mitigate FIs at runtime. Figure 3 below presents simplified relations among the terms and concepts defined in ISO 26262 and SOTIF.

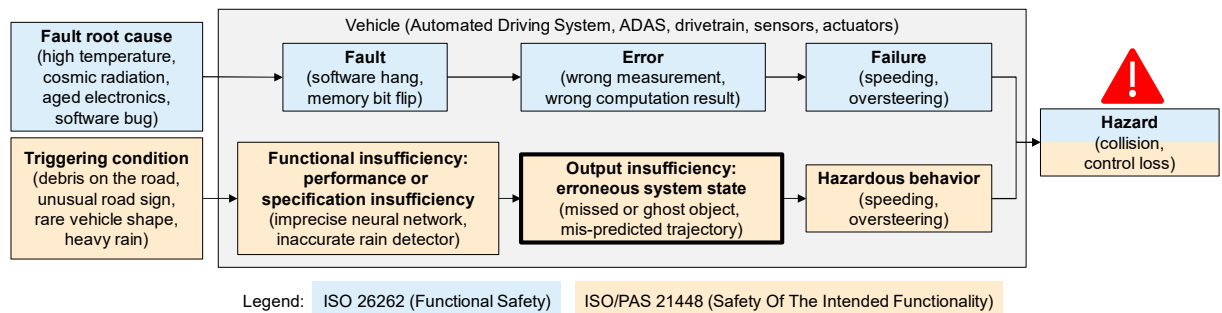

**Figure 3. Simplified cause and effect hazard models in ISO 26262 (in blue above) and SOTIF (in yellow below). The grey *Vehicle* box separates what happens inside and outside of the vehicle. Examples are given in brackets. The bold *Output insufficiencies* box identifies the focus of our classification.**

FIs in autonomous vehicles have been logged and studied both in real-life road tests and in academic research. For example, a number of real-life consequences of FIs are registered by the California Department of Motor Vehicles (DMV). DMV publishes AV collision reports and disengagement reports from a wide range of AV manufacturers every year [18, 19], in which system failures as well as FIs are reported.

Some statistical analyses have been conducted based on the DMV reports: [20] provides a processed dataset of the DMV reports from 2014 to early 2020; [21] analyzes frequencies rather than causes of AV disengagements and accidents; [22] finds that most disengagements are related to machine-learning-based perception and decision-and-control systems. When categorizing the causes of the disengagements and accidents, however, [22] does not clearly separate faults and FIs. While system faults are much better studied, FIs are less systematically addressed at design time and hence have to be handled at runtime. To focus ADS improvements on biggest issues first, the automotive industry needs classification and prioritization of the FIs contributing to dangerous situations.

There is also research on FIs based on pre-recorded sensor datasets and environment conditions simulated by AD simulators. For example, [23] proposes a methodology on discovering novel triggering conditions for perception FIs with a case study of a LiDAR-based AD perception system; [24] proposes a workflow to identify triggering conditions defined in [8] for LiDAR and camera-based perception module and evaluates the proposal with simulation tools.

To our best knowledge, no detailed classification of FIs of ADS internals have been made in existing studies of FIs and most of the prior works focus on the triggering conditions of FIs. In this paper, we do not study and categorize the triggering conditions, but focus on FIs in the perspective of SOTIF, and distinguish them from systematic failures and errors, which have been well studied in the scope of ISO 26262. Our comprehensive characterization of FIs will be helpful in the design of mitigation mechanisms for ADS FIs as well as in triggering conditions analysis.

## MITIGATION OF FUNCTIONAL INSUFFICIENCIES

In this subsection we discuss the existing FI mitigation strategies. One mitigation strategy is to reduce ADS FIs by avoiding the external triggering conditions using restricted Operational Design Domains (ODD) [25, 26]. Alternatively, we can limit the frequency of FIs occurrence by improving the performance of ADS at the component and system levels.

At the component level, FIs can be mitigated by improving the performance of a specific ADS module or using an alternative method. For example, ADS perception can be improved by more accurate sensors; an FI in a lane detection algorithm can be mitigated by exploiting high-definition maps and enhanced localization algorithms [8]; ghost objects can be decreased by using a different detection mechanism; missed objects can be reduced by combing the radar and the vision system. Similar approaches can also be applied to motion planning and actuation algorithms [27]. [10] details the pros and cons of different sensors and sensor fusion algorithms. [16] analyzes FIs of two commonly used sensor fusion methods and provides several suggestions on how to improve the multi-sensor fusion algorithms. A similarity-based incremental learning algorithm is proposed in [28] to improve the learning model for pedestrian prediction over time. [15] presents an extensive survey of existing human motion prediction algorithms and suggests that a combination of multiple prediction algorithms could be the approach to a more robust prediction in unknown situation. Besides applications in the AD prediction module, neural networks are also widely used in the perception module to detect and identify objects. [29] shows that the ensemble performance of multiple bitwise neural networks can surpass the performance of a single high-precision neural network in terms of classification accuracy. The Responsibility-Sensitive Safety (RSS) [30] and the Safety Force Field (SFF) [31] algorithms check if the planned trajectory is compliant with the ADS safety constrains and predefined traffic rules, so as to prevent the AV from being the cause of a road accident from the legal perspective. These algorithms are able to mitigate some hazardous situations caused by FIs in the ADS motion planner without actively detecting or identifying the FIs, because only actuation control commands that are within the safety envelope of the ADS are passed on by the RSS or the SFF modules to the AD actuators. Similarly, [8] mentions applying restrictions on AD functions as a way to prevent or mitigate SOTIF-related risk. However, as [13] points out, such approach may compromise the usability of the ADS.

At the system-level, FI can be addressed by monitoring, redundancy, diversity (heterogeneity), functional restrictions or other measures [8]. The popular monitor/actuator design pattern from [27] enables shutting down the main driving channel in a fail-silent way, if it violates safety criteria according to the monitor. [32] proposes a set of application-specific verification methods to ensure the correct arbitration in a fault-tolerant redundant AD architecture with two heterogeneous AD channels. However, [32] emphasizes that the diversity in AD channels will result in inconsistent output due to the problem of replica indeterminism [12]. As a result, the traditional voting approach such as majority voting or triple modular redundancy (TMR) among the redundant channels is not suitable. To cope with the replica indeterminism, more sophisticated comparison techniques, weighing mechanisms, etc. need to be deployed in the heterogeneous ADS. [33] proposes a Level 3 ADS to mitigate exiting ODD [25, 26], which is a safety-critical FI, by transferring the control to the fallback human driver. [13] proposes an AD architecture where the safety channel and the nominal channel monitor the health status of the other. The safety channel in [13] has its own set of sensors and a simplified world model. Although primarily aiming at mitigating faults and failures in the ADS, [13] mentions that the safety channel might be able to help address certain FIs due to its heterogenous implementation. However, both proposals from [33] and [13] can cause dangerous situations due to possibly insufficient time budget for the human driver to take over [34]. [11] proposed a commander/monitor dual-channel AD architecture to reduce false positives in free space detection. In their proposal the free space used for motion planning is the intersection of free space identified by the commander channel and the monitor channel. In other words, the common free space detected by both AD channels.

Companies in the automotive industry proposed several architectures to address the ADS FIs on the system level as well. Mobileye's True Redundancy [35] ADS uses two independent channels to reduce FIs in AD perception. One channel of the True Redundancy ADS uses only cameras for perception, the other uses only radars and LiDARs. The sensors in different channels do not interact with each other and each channel builds an independent world model. Instead of focusing only on improving the perception module performance, a Safety Shell algorithm is proposed in [36] to calculate the risk of AD channels based on the cross-channel analysis of the world models and trajectories. An arbitration logic is then applied based on the channel last safe intervention time derived from the risk calculation [36]. Instead of a specific algorithm or implementation proposal, our work proposes a generic design pattern to address FIs in ADS based on our FI characterization and experimental results on our open-loop simulation setup with realistic AD frameworks.

From the above overview, we conclude the following. The existing works tend to mix faults and FIs when discussing the improvement proposals. An accurate and extensive classification of FIs is currently missing in the literature. In addition, there are no clear winning design patterns for ADS to handle FIs at the system-level yet. Therefore, we would like to propose a clear characterization and categorization of FIs and a system-level approach to address FIs.

# 3. CHARACTERIZATION OF OUTPUT INSUFFICIENCIES

This section first motivates for characterization of OIs, then describes the method of video-based analysis of existing ADS in real traffic scenarios to characterize FIs. Finally, the Results subsection presents a table with OI characteristics.

## MOTIVATION TO FOCUS CHARACTERIZATION ON OUTPUT INSUFFICINCIES

SOTIF divides FIs (functional insufficiencies) into performance and specification insufficiencies, which can be diverse in nature and origin and, consequently, hard to classify in a comprehensive way. On the other hand, FIs result in OIs (output insufficiencies), as shown in Figure 3. OIs can be easily attributed to the few major internal ADS functions, such as perception or path planning. Moreover, analyzing ADS FI root causes (e.g. triggering conditions or design property) requires detailed information about all internal hardware and software components, which practically complicates the classification of FIs if the ADS is studied as a black box. Therefore, we focused our characterization on OIs, which should cover most of the FIs.

## CHARACTERIZATION METHODOLOGY

The methodology of our characterization study is to inspect the DMV reports and public road test videos to enumerate, classify and characterize FIs and to assess their prevalence and severity in automated driving.

First, we performed statistical analysis of over 2500 disengagement registered in the 2021 DMV disengagement report [19] to learn about reasons of disengagements in real-life cases. We classified the disengagement causes and the frequencies of their occurrence. The statistical analysis helped us gain knowledge of the most frequent contributing reasons of disengagement. The details of our DMV report study are presented in the Characterization results subsection below.

While the DMV reports provide a large amount of data, the text reports are often ambiguous because different companies provide different levels of details in the description of facts causing disengagement [19]. Fortunately, recent advances in prototyped ADS also resulted in many public video recordings of autonomous vehicles driving on public roads with real diverse traffic. Therefore we studied such video recordings as well to obtain a more intuitive understanding of the disengagement causes and consequences. Remarkably, the videos often include display screens with environment perception and motion plan of the ego vehicle, see an example in Figure 4. Several recordings even deploy drones [2] to accompany the in-cabin videos with a birds-eye-view perspective on the road and traffic.

By comparing the real-life video of the vehicle surroundings with the ADS state on the display, an attentive human observer can spot differences between the physical environment around the vehicle and the world model the ADS perceives, which are OIs of the ADS. For example, the figure below shows the real-life view with a traffic light pole that is meanwhile absent on the ADS display. To compensate for the ADS insufficiencies the driver in the video recordings sometimes takes over control of the vehicle in order to ensure safe driving. By extrapolating such situations as if no disengagement took place, the human observer of the videos can judge the severity of the ADS output insufficiency in the traffic scenario. Note, that such insufficiencies are not caused by system faults, as faults are often properly detected by the ADS safety monitors resulting in a warning to the driver or other fault handling mechanisms.

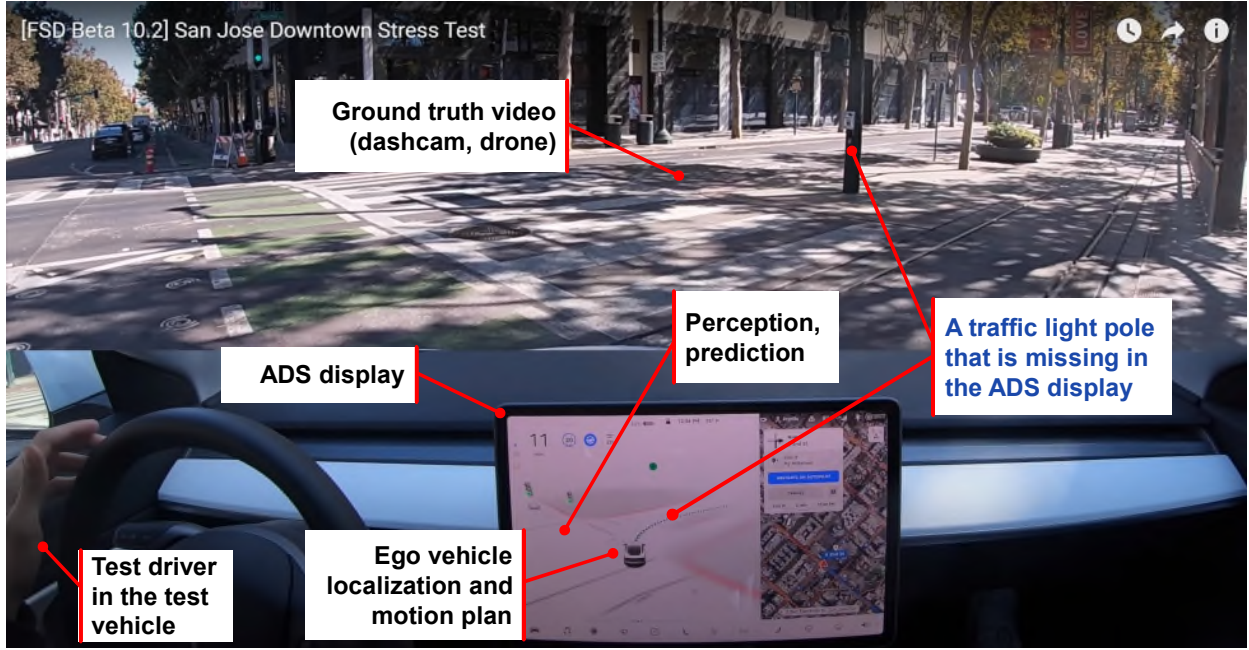

Figure 4. Road test video screenshot with a ground truth view above and an ADS display below identifying an insufficiency.

## CHARACTERIZATION RESULTS

The distribution of causes of the disengagement from the 2021 DMV disengagement report [19] is depicted in Figure 5 below. Note that the "out of scope" category means the disengagements were correctly and automatically triggered by the ADS without leading to any hazardous situation; "fault" category includes all the reports that clearly claimed that a software or hardware or other systematic faults had occurred. Figure 5 shows that over 90% of the DMV report data is analyzable for our purpose and obviously disengagements caused by insufficiencies occurred five times more than those caused by system faults.

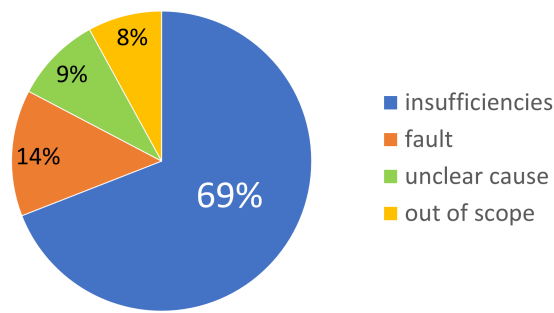

Figure 5. Causes of disengagement in 2021 DMV disengagement report.

Videos of different road test scenarios we studied are publicly available, such as [2], [3], [5], [9]. An overview of the reviewed videos is summarized in Table 1. In those videos, the ADS did not warn the driver about internal systematic faults nor showed any sign of internal systematic faults when the insufficiencies were observed. Notably, as shown in Table 1, 87% of the insufficiencies we registered could have resulted in a collision or a hazardous situation on the road, should the driver had not intervened with AD disengagement, or other road users had behaved differently.

**Table 1. Summary of studied videos of road tests.**

| Number of watched videos | 32 |
|---|---|
| Watched hours | 10 hours 17 minutes |
| Total driving scenarios with OIs | 71 (100%) |
| Safety-critical scenarios with OIs | 62 (87%) |
| Number of ADS brands | 9 |

Based on our DMV report analysis and the video observations, we concluded that ADS exhibits OIs against one of the three criterion: ground truth, human intuition and traffic rules. For example, a missed object by the perception system is an ADS deficiency against the ground truth; the inability to reduce the speed on urban roads is a OI against the traffic rule; the too conservative motion plan of an ADS is against the driving style based on human intuition.

We also distinguish 16 OI types and divided them into four categories: world model, traffic rule, motion plan, and ODD as listed in Table 2 with their characteristics. The categories identify the ADS subsystem responsible for the OI. The OI names describe the ADS insufficiency. For example, insufficiency #1 identifies the system's inability to find the location of the ego-vehicle. The reference column specifies what the insufficiency is relative to. The typical timing nature of the insufficiency is captured in the timing column. OIs can exist either for a long time spanning several seconds or sporadic. For example, insufficiency #2 can exist throughout the ADS operation time, while insufficiencies #8 and #11 usually only exist for a short time and refer to the predicted motion in future time. Finally, the sensors column gives examples of sensor modules involved in triggering the insufficiency. Note that several OIs do not involve sensors at all, such as insufficiencies #10 or #13.

**Table 2. Output insufficiencies classification and characterization.**

| ID | category | name | criterion | ADS module | sensors | timing |
|---|---|---|---|---|---|---|
| 1 | world model | wrong ego-vehicle localization | ground truth | localization | GNSS, IMU, lidar, camera | sporadic, long |
| 2 | world model | wrong map | ground truth | map | HD map files | long |
| 3 | world model | missed object | ground truth | perception | lidar, radar, camera, ultrasonic sensor, microphone | sporadic, long |
| 4 | world model | ghost object | ground truth | perception | lidar, radar, camera, ultrasonic sensor, microphone | sporadic, long |
| 5 | world model | wrong object position, orientation, or dimension | ground truth | perception | lidar, radar, camera, ultrasonic sensor, microphone | sporadic, long |
| 6 | world model | wrong object classification | ground truth | perception | lidar, radar, camera, ultrasonic sensor, microphone | sporadic, long |
| 7 | world model | wrong drivable space identification | ground truth | perception | lidar, radar, camera, ultrasonic sensor | sporadic, long |
| 8 | world model | wrong object trajectory | ground truth | prediction | - | future, sporadic |
| 9 | traffic rule | wrong traffic sign, light, lane marking or operator recognition | traffic rule | perception | camera, V2X | sporadic, long |
| 10 | traffic rule | violation of traffic regulation (e.g. right of way) | traffic rule | motion planning | - | sporadic, long |

| 11 | motion plan | counter-intuitive motion plan | human intuition | motion planning | - | future, sporadic |
| --- | --- | --- | --- | --- | --- | --- |
| 12 | motion plan | indeterminate motion plan | human intuition | motion planning | - | sporadic |
| 13 | motion plan | unsafe planned trajectory | human intuition | motion planning | - | sporadic, long |
| 14 | ODD | wrong weather classification | ground truth | ODD checker | rain and light sensor, visibility range sensor | long |
| 15 | ODD | wrong road classification | ground truth | ODD checker | road surface sensors, GNSS | long |
| 16 | ODD | wrong traffic classification | ground truth | ODD checker | camera, radar, clock, GNSS, V2X | long |

Each OI from Table 2 can lead to hazardous situations. Besides obviously dangerous OIs such as #3 "missed object", seemingly innocent OIs, such as ODD- or motion planning-related ones, can trigger a sequence of events leading to a hazard. For example, due to a wrong weather classification categorized by insufficiency #14, the camera usage can be allowed in adverse light conditions that can lead to an accident. A counter-intuitive motion plan #11 may confuse other traffic road users and cause a collision due to misunderstanding in motion negotiations.

Figure 6 illustrates the distribution of OIs per OI category as defined in Table 2 from both the 2021 DMV disengagement report and the road tests we studied. The distribution patterns are similar in both DMV report and the real-life video study. The plot also suggests that OIs related to the world model category and the motion plan category are the most prevalent.

In summary, in our studies we observed only a limited number of system faults and, yet, modern ADS often suffered from many OIs that could lead to severe hazards. Therefore, to enable focused ADS improvements, knowledge of the most frequent contributing OIs to dangerous situations are needed.

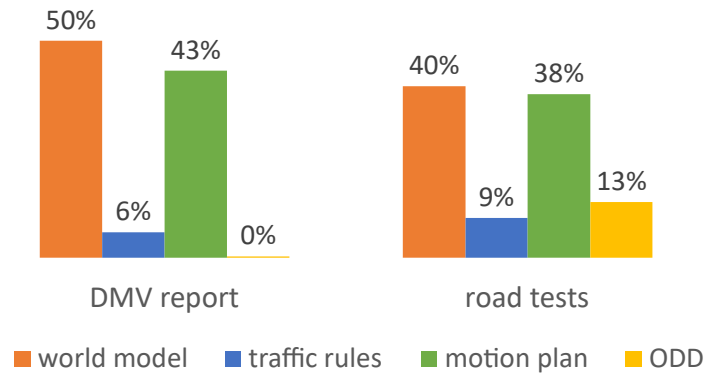

Figure 6. Distribution of functional insufficiencies in the 2021 DMV disengagement report and the road tests we studied.

# 4. MITIGATION OF FUNCTIONAL INSUFFICIENCIES

This section experimentally defines an architectural design pattern to mitigate FIs.

## HYPOTHESIS ABOUT HETEROGENEOUS AD CHANNELS AND FUNCTIONAL INSUFFICIENCIES

Redundancy is commonly implemented by car manufacturers to improve the vehicle safety. For example, BMW includes three AD channels in their ADS [7]. To avoid systematic faults and common cause failures [6], heterogenous channels do not reuse hardware, software, algorithms or, possibly, sensors. AD systems individually developed by different car manufacturers can be seen as heterogenous AD channels. Interestingly, in our OI characterization work described in Section 3 we noticed that the distribution and frequency of OIs reported by different car companies varied from each other. We therefore make the following hypothesis:

*There exist times when some AD channels in a heterogenous multi-channel ADS encounter functional insufficiencies, while the other AD channels do not.*

A motivation scenario is shown in Table 3 below to illustrate this hypothesis. In this example we assume that all AD channels in the ADS can have FIs at certain times during the driving. For instance, at time steps 3, 5, and 7 one or two channels have FIs but there is still at least one available channel that does not encounter FI. Therefore, if the limited channel is in control of the vehicle at that time, it can switch to a channel not affected by FIs, instead of issuing disengagement or performing an emergency brake. The ADS can then stay in the AD mode and continue driving. In this way the ADS safety and availability are enhanced. If this hypothesis is true, there are opportunities to switch between complementary channels at runtime to mitigate FIs on top of existing implemented methods inside each AD channel, such as sensor fusion. Furthermore, at time step 4, although FIs are triggered in all channels, there still exist possibilities for the ADS to switch to the channel that does not have safety-critical FIs based on the evaluation of the current states of all the AD channels or the ADS can perform an evasive maneuver. Such AD channel evaluation can be done, for example, via the risk and arbitration calculation proposed in [36].

Table 3. A motivation scenario showing functional insufficiencies in different AD channels of the ADS at different time steps.

| Time step | 1 | 2 | 3 | 4 | 5 | 6 | 7 |
|---|---|---|---|---|---|---|---|
| Channel1 | OK | OK | OK | FI | FI | OK | OK |
| Channel2 | OK | OK | OK | FI | OK | OK | FI |
| Channel3 | OK | OK | FI | FI | OK | OK | FI |

## OPEN-LOOP SIMULATION METHODOLOGY TO VERIFY THE HYPOTHESIS

Ideally, we would like to validate our hypothesis by experimenting with multiple heterogenous AD channels operating simultaneously in the same test vehicle. However, we do not have access to such a test vehicle. The available DMV reports and the road test videos do not allow us to evaluating different ADS in the exactly same driving scenario either. First, the multiple AD channels in the test vehicle are not transparent in both the DMV reports and the videos; second, in some videos even when driving multiple times the same vehicle along the same route, the test vehicle was exposed to different environments (such as different lighting and traffic). Therefore, we set up an LG SVL simulation environment for autonomous vehicles [37] with three realistic AD channels: Baidu Apollo 5.0 [38, 39], Autoware.Auto AVP [40], and Comma.AI openpilot [41]. In the simulation environment we can compare the AD channels against the "ground truth" of the simulator, see Figure 7 with screenshots of the simulator and AD channel GUIs. The screenshots of the GUIs are taken at the same simulation moment, when a cyclist is about to cross the ego vehicle's trajectory.

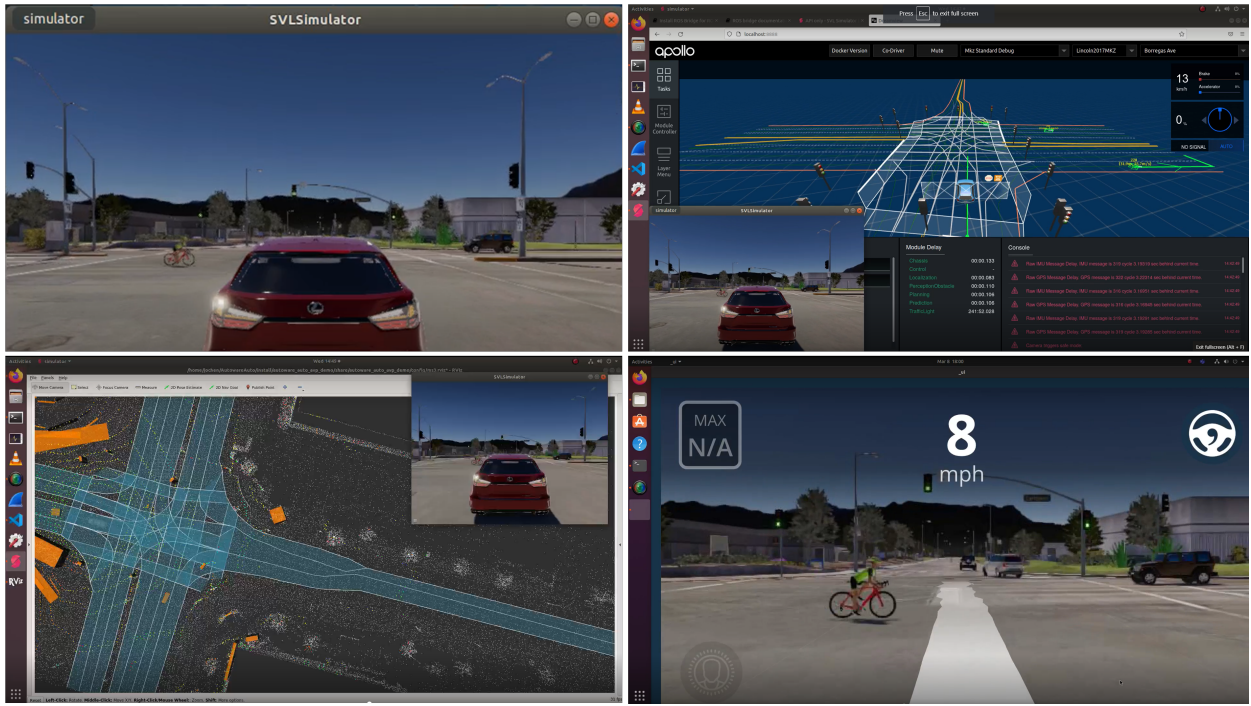

**Figure 7. Screenshots of the LG SVL simulator (upper left), Apollo 5.0 Dreamview GUI (upper right), Autoware.auto Rviz GUI (lower left), openpilot GUI (lower right) at the same simulation moment.**

The AD channels we chose for our simulation study, despite having different capabilities, have been used to drive real-life vehicles autonomously. Ideally, we would like to have three AD channels running in parallel in the same simulation run. However, due to high engineering complexity and high hardware requirements we resorted to running them sequentially. To ensure that the environment is exactly the same for every undertaken experiment, we employed an open-loop simulation where the AD channel's control over the vehicle was disabled and the ego-vehicle was driven by the same script asserting a fixed throttle (apply_control() with sticky=True in the LG SVL simulator). Furthermore, the traffic was controlled through a fixed random number generator seed, assuring that each scenario would show exactly the same traffic objects and motions. Finally, we used the same vehicle model in the LG SVL simulator but equipped it with various sensors required for the different AD channels. Figure 8 illustrates our experimental open-loop simulation setup.

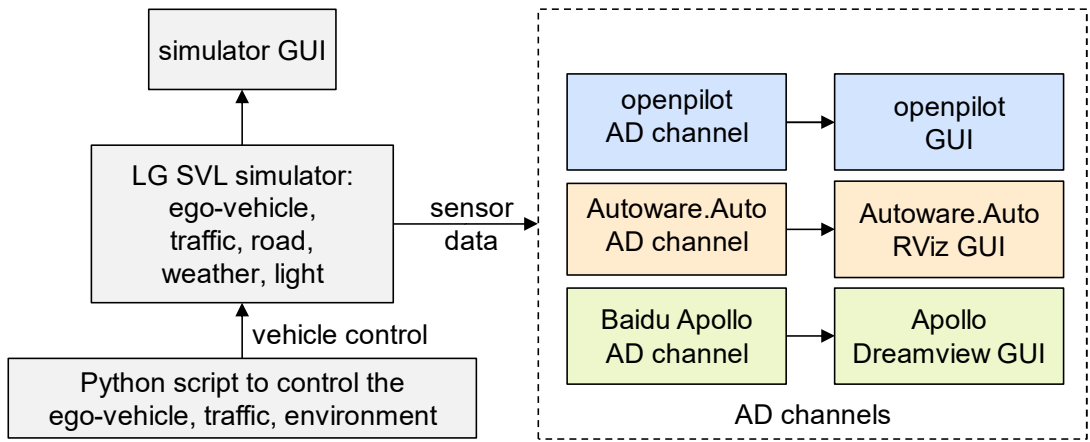

**Figure 8. Open-loop simulation setup with sensors feeding AD channels and a Python script controlling the ego-vehicle and the traffic.**

Below is an overview of our simulation platform and scenarios with Table 4 contrasting the AD channels we used:

- Alienware Aurora workstation, Intel Core i7-8700 and Nvidia GeForce GTX 1070, 64GiB SDRAM
- Ubuntu 18.04, desktop
- LG SVL simulator version 2021.3, the Borregas Avenue crossroad map
- Driving scenarios:
    - the ego vehicle drives straight with a fixed acceleration from a fixed starting position to a fixed end position using a Python script
    - 10 seeds for random crossing traffic and 1 scenario with one vehicle in front of the ego vehicle performing a left turn
    - all traffic lights are configured to be always green
    - good lighting and clear weather (no rain, fog, wetness, nor cloudiness), 12:00am

Table 4. Overview of the three AD channels used in the experiments.

| Channel name | Version/tag | Sensors | Key algorithms & features | ODD |
|---|---|---|---|---|
| Baidu Apollo | 5.0 | lidar, camera, radar, GNSS, IMU | Deep neural networks, various computer vision algorithms, Kalman filter, configurable high-level sensor fusion, HD map | Urban |
| Autoware.Auto | AVP demo | lidar, GNSS, IMU | Euclidian cluster, NDT matching, HD map | Parking |
| openpilot | 0.8.10 | camera, (radar not used) | Deep neural network, MPC, Kalman filter | Highway |

It is worth mention that the publicly documented LG SVL simulator setup did not use the advanced camera-radar-lidar fusion in Apollo 5.0 and relied purely on the lidar for object detection. In our experiments, we also manually created derivative Apollo channels by creating new configurations to fuse lidar and camera sensor data in Apollo 5.0. But we didn't observe any performance improvements in the Apollo visualization GUI. Also the openpilot was not utilizing the radar sensor in our simulations.

Besides the limited features implemented in the chosen AD channels, the hardware computational and memory resources can also be a triggering condition that causes FIs. In particular, we observed in our experiments that if the GPU power was not sufficient to run the simulator and process the AD channels, a video frame could be dropped resulting in one or more missed objects. Therefore, for demanding use cases we ran the simulator and visualization on a machine separate from the AD channel machine to keep the GPU processor utilization lower than 60% and video memory usage of less than 4GiB. Such hardware considerations must also be included in the analysis of FIs next to external triggering conditions as described in [8]. For example, at a busy intersection with many road users, the computational and memory limits of the hardware can lead to degraded performance of the ADS and, subsequently, to hazards.

## EXPERIMENTAL RESULTS

Our open-loop simulations revealed 42 OIs, see Table 5. The result shows that first, some OIs from Table 2 are absent in our simulations due to the limited number of simulated driving scenarios; second, the AD channels have different capabilities which severely influence their OIs distribution per category in Table 5. For example, in our experiments Apollo is the only channel with visualized predictions of road users, and the Autoware.Auto and openpilot channels do not analyze traffic lights, because they are outside of their ODDs. Note that we therefore do not perform the imbalanced comparison between the channels. For example, for openpilot we only counted the objects on the ego vehicle's trajectory and excluded detection of objects outside of the camera field of view. Nevertheless, we can conclude that even the most capable channel, which is Apollo in our experiments, has limitations. Finally, Table 5 clearly suggests that different AD systems have different types of OIs at different frequencies, e.g. Apollo had only 2 missed objects, while Autoware.Auto and openpilot had twice as much.

Table 5. Output insufficiencies of the AD channels identified in the open-loop simulations.

| Functional insufficiency | Apollo | Autoware.Auto | openpilot | Total |
|---|---|---|---|---|
| wrong ego-vehicle localization | - | 2 | - | 2 |
| missed object | 2 | 4 | 17 | 23 |
| ghost object | - | 1 | 3 | 4 |
| wrong object position, orientation, or dimension | 2 | 1 | - | 3 |
| wrong object classification | 1 | 1 | - | 2 |
| wrong drivable space identification | - | - | 4 | 4 |
| wrong object trajectory | 4 | - | - | 4 |

The goal of our simulations was to validate the hypothesis that there exist times where not all heterogeneous channels have OIs that lead to a hazardous situation. In general, we observed that the most capable Apollo channel performed very well in all of our simulations and could have been the best channel to drive the vehicle for most of the time. Nevertheless, we spotted several cases when even the best channel encountered insufficiencies, while the others did not:

1. In one random traffic simulation a cyclist was quickly crossing the trajectory of the ego-vehicle, see the illustrations of the AD channels in Figure 7. Apollo detected the bicycle only after it crossed the ego-vehicles route, while Autoware.Auto was able to trace the cyclist earlier, as soon as it reached the crossroad. Consequently, Apollo had a false negative detection in the perception system, which we term "#3 missed object" in our classification from Table 2. Note that false negatives may lead to severe accidents.
2. In a scenario with a single vehicle in front, the Apollo channel overestimates the front vehicle dimensions (insufficiency #5 in Table 2), which can result in sudden braking or evasive maneuvers that are unexpected to the rest of the road users. Such false positives may be less severe than false negatives, but may sacrifice automated driving availability and comfort. In contrast, the Autoware.Auto perception system correctly estimated the size of the front vehicle.
3. In another random traffic simulation Apollo predicted that an approaching truck from the opposite direction would cross the ego vehicle's way (insufficiency #8 in Table 2). Such false positive predictions can cause the ego-vehicle to perform unnecessary emergency braking.

Conclusions from our open-loop simulations are illustrated in Figure 9 in a qualitative way using a Venn diagram based on [36] and the SOTIF classification of driving scenarios into four areas in SOTIF Section 4 [8]. The three channels in Figure 9 have different capabilities, such as traffic light recognition or road user motion prediction, which are identified with colored rectangles. Circles represent FIs of each channel as a hole in its capability. Despite having larger capabilities, FIs still exist in advanced AD channels. And the (sometimes smaller) FIs of the advanced channels could be covered by capabilities of the other less advanced channels. In other words, capabilities of some (less advanced) AD channels can be complementary to the other (more advanced) channels. Therefore, in Figure 9 the shared insufficiencies, which are identified by the white hole, represent the same FIs that exist and will be triggered by the same triggering conditions in all the channels. These shared FIs are not covered by the capabilities of any channel in the same ADS. However, the shared FIs are smaller than any of the total FIs of each individual channel (represented by the big circle in each channel). Noteworthy, the common capabilities of the combined multi-channel system can be potentially bigger than that of the most advanced channel alone. The white hole in AD capabilities closely relates to the unknown hazardous scenarios in the SOTIF classification.

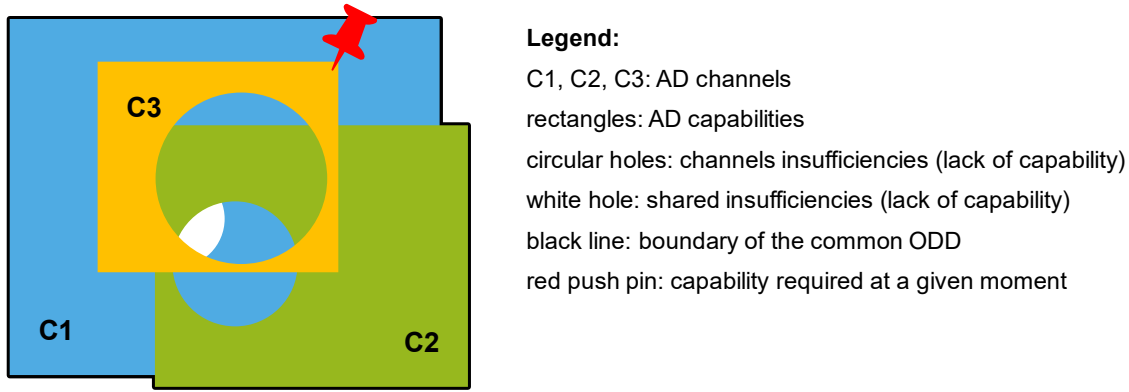

**Figure 9. Conceptual illustration of overlapping functional insufficiencies and capabilities of multiple AD channels. Capabilities of the channels can be made complementary to enable runtime selection of the channel without insufficiency.**

At runtime, the driving situation changes constantly, which can be represented by an imaginary push pin moving on Figure 9 and the pinpoint specifies the currently required capabilities for safe continuous driving. If the pin ends up on a colored block, there is at least one AD channel that has sufficient capability to drive safely under such situation. Note, that the ADS always has to select a channel that has a color dot under the pin. For example, if the pin ends up on the upper right corner of channel C3, as shown in Figure 9, then the ADS should either choose channel C1 or C3, but not channel C2. However, if the pin lands inside the white hole or even in the area outside of the black border line (for example, when the ADS is outside of ODD), no channel has the required capability and the ADS will suffer from the shared insufficiency existing in all the channels. In such cases the ADS has to perform an emergency brake or disengage from the AD mode to hand over control to the human. Just like any AD channel will have certain FIs, the shared FIs are also inevitable. But by using multiple heterogeneous AD channels, we believe it is possible to minimize the shared FIs within the ODD. From the SOTIF perspective, the design time goal is to construct an AD system, which has a minimum white hole of capabilities inside its ODD and which can switch between channels at runtime to choose the most capable channel for the current driving scenario.

In order to mitigate FIs, the multi-channel ADS has to continuously monitor heterogeneous AD channels at runtime and switch to a more capable AD channel when a FI occurs in the current AD channel. Based on this observation in the next section we propose an architectural design pattern used to select the least limited channel at runtime. We named our design pattern Daruma.

## DARUMA: ARCHITECTURAL DESIGN PATTERN FOR CROSS-CHANNEL ANALYSIS AND ARBITRATION

Figure 10 below presents our Daruma architectural design pattern for multi-channel ADS. The Daruma design pattern leverages cross-channel analysis to dynamically select a channel that has sufficient capabilities to safely and comfortably continue driving the vehicle in the current situation. The traditional ADS components are shown in grey boxes in Figure 9; the new or modified components are identified as the colored boxes and italics text. The heterogeneous AD channels (channel1 to channelN) include AD functions and (possibly shared) sensors depicted in Figure 1. The output of the Daruma architecture is low-level actuation setpoints that are sent to the vehicle actuators. The monitoring techniques from prior works are part of either the AD channels or the components shown as "fault monitors, ODD checker, ..." in Figure 10. Traditionally, the arbitration decision of the multi-channel ADS is based on the output of intra-channel safety mechanisms, platform fault monitors and other metrics such as comfort, efficiency and availability. In general, the high-level arbiter is similar to the Mode Manager from [42] and controls the selection of the driving channel. However, the high-level arbiter in the Daruma architecture also receives output from the cross-channel analysis, which can trigger selection of a different (safer) channel even in the absence of faults.

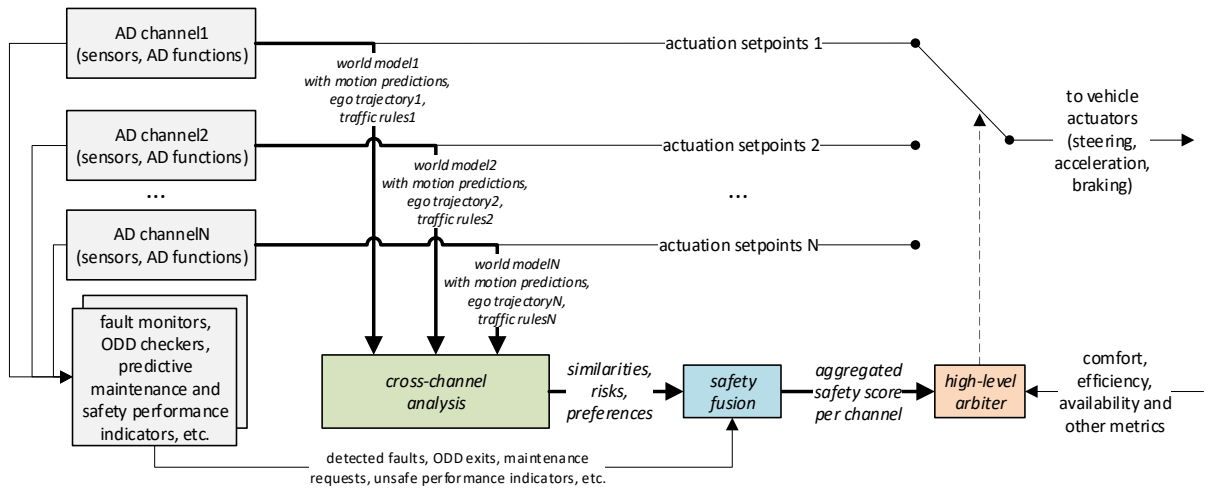

**Figure 10. Architectural design pattern with cross-channel analysis and AD channel arbitration to mitigate functional insufficiencies in ADS.**

To perform cross-channel analysis, Daruma architecture taps additional information from the AD channels, including the world models, object motion predictions, ego vehicle motion plans and traffic rule assessments. This high-level channel state information is often shown on ADS displays in the autonomous vehicles to inform the passenger about vehicle's internal state and planned actions.

Cross-channel analysis can involve several state and design properties of the multi-channel system:

1. *Similarities* of channels' high-level states. If multiple channels identify an object to be present in the same location with a similar motion plan, there is a higher chance of finding this object in reality, provided that the channels are sufficiently heterogeneous and reliable. Each color in Figure 11 represents output from one channel. Obviously, the three channels output different object locations, shapes and trajectories. Major differences among these channels are:
    a) the green channel doesn't see the vehicle behind the ego,
    b) the child crossing the street is only observed by the red and blue channels,
    c) the red channel plans to stop, the green and the blue channels plan to continue driving straight.
2. Cross-channel *risk* of collision, driving off-road, or losing control over the vehicle. The new cross-channel analysis component performs cross-checking of ego vehicle motion plans against the world model and object predictions generated in each channel. Therefore, for 3 channels we can compute risk as a function of future time for 3 ego motion plans times 3 world models with a total of 9 risk functions, see Figure 11. Even violating the traffic rules can be expressed as a safety risk. For example, one channel outputs an ego stop trajectory, while the other requests an acceleration of the ego vehicle. From the availability perspective, the second channel will be preferred. However, if the first channel complements the ego vehicle stopping trajectory with a red traffic light annotation, the second channel can get a low safety score in the end.
3. Design-time or runtime channel *preferences*. The preference can be expressed in weights to promote the most advanced channel for the current ODD or in terms of advance switch times as described in [36].

Straightforward majority voting using similarities of channels may not yield the safe arbitration. For example, according to difference b) the green and blue channels agree, while only the red channel is correct about the pedestrian in front. Therefore, inter-channel similarity alone as a metric is not sufficient to judge the situation and select the optimal trajectory. On the other hand, similarity can be captured by matching risk profiles from cross-channel analysis or using credit-based temporal matching, where the confidence in having a match increases if the match occurs several times in a row.

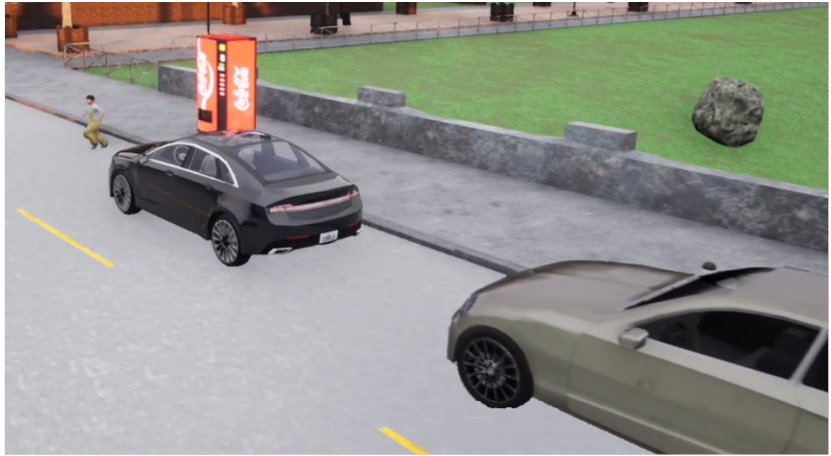
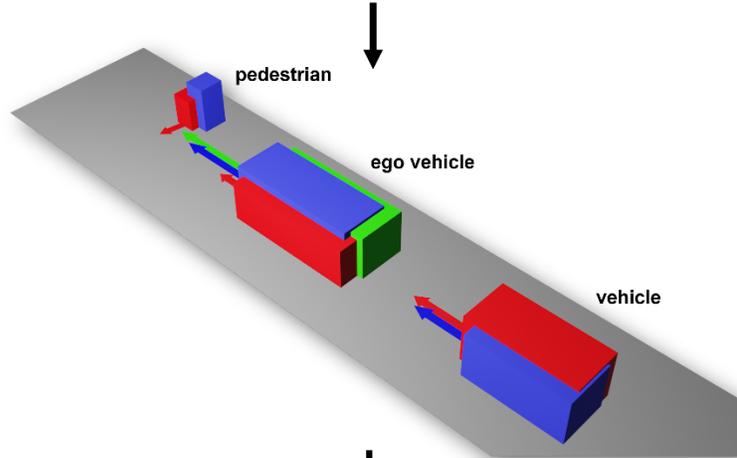
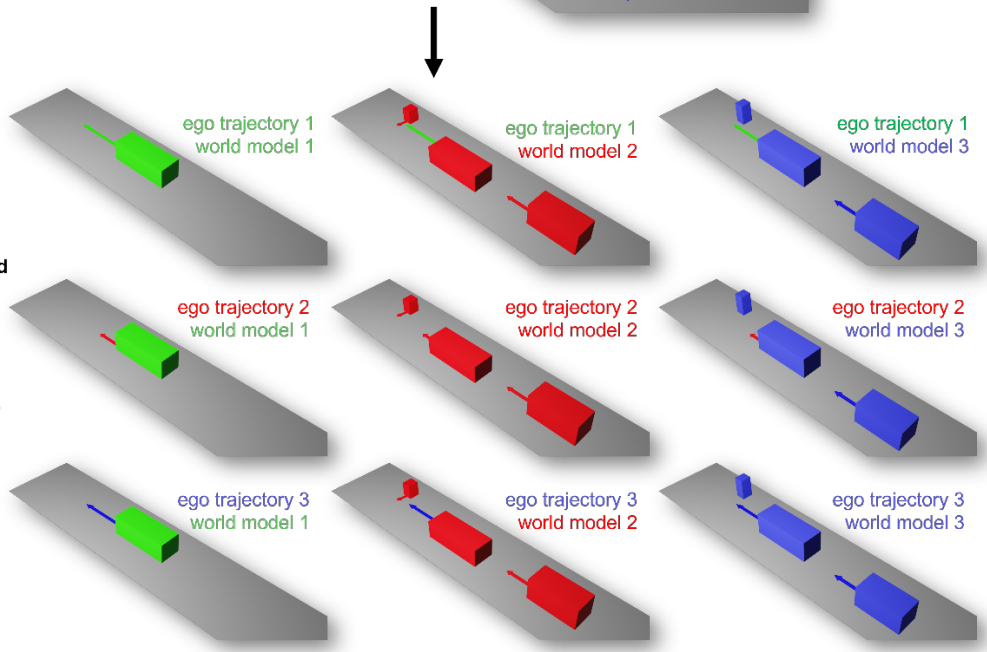

Figure 11. Cross-channel analysis by using juxtaposition of high-level states of three AD channels. The driving scenario on the top is processed by the red, green and blue channels to compute high-level channel states. Then, by applying the Daruma design pattern, the ego vehicle trajectories from each channel are cross-analyzed for hazards in all world models forming a 3x3 matrix of possible driving scenarios.

Algorithmic choices for cross-channel analysis include:

1. *Geometric overlay* of the high-level channel current and predicted states, which is partially used in [32], focuses on detection of collisions and off-road driving based on geometric intersections of 2D or 3D shapes. Remarkably, it avoids AD disengagements when the AD channels differ in a safe way, which improves ADS availability. For example, if the one channel suggests to avoid obstacle on the left and the other – on the right, then the geometric overlay will show that both channels are safe although they disagree.
2. *Risk-based calculation,* where the motion plans from one channel are analyzed against world models and object motion predictions from other channels. This cross-channel analysis computes the probability of collision in future, e.g. using the last safe intervention time [36] to identify a last time moment to switch to another AD channel or to perform an evasive maneuver on time. Note that this approach greatly improves availability of ADS by giving channels the time to correct themselves and filter out intermittent issues.
3. Another interesting method for the cross-channel analysis is *machine learning*. For example, [43] proposes neural network RiskNet to identify collision risks in camera images using optical flow. By analogy, one can define a neural network model operating on the high-level cross-channel state information or its derivatives.

In the Daruma design pattern the inter-channel and cross-channel analysis results (e.g. similarity matches, risk functions, preference weights) are fed into the safety fusion module. The safety fusion module computes an aggregated safety score per AD channel based on the cross-channel analysis results and the traditional metrics. Subsequently, the aggregated safety score allows the high-level arbiter to select a safe channel while taking into account comfort, efficiency, and other metrics. Various techniques can be applied to the safety fusion, such as hierarchical decision tree or a numerical weights calculation. Note that in a simple implementation of the safety fusion, the aggregated safety scores can be binary – either zero or one to clearly indicate preference of the safety subsystem.

The effectiveness of the proposed Daruma design pattern strongly depends on the quality of the AD channels. If a channel is poor in many situations, it can decrease the overall system performance. Therefore, it is important to develop and select diverse channels with complementary capabilities and overall high performance at design time. In other words, the designer has to reduce the white hole in Figure 9 by properly dimensioning and positioning the channel capabilities.

The proposed Daruma architectural design pattern enables advanced cross-channel analysis techniques *without compromising AD availability*. Instead of engaging in a traditional safe stop or disengagement upon detection of disagreement among channels, Daruma architecture enables switching to a safer channel to continue safe driving with minimum discomfort. Another important virtue of the proposed design pattern is scalability. Due to its modular design the number of channels in Daruma can be increased to boost safety without compromising AD availability. Compared to the monitor/actuator approach described in [27], the scalability of Daruma is an important advantage to create additional commercial value. Moreover, the Daruma architectural design pattern is compatible with state-of-the-art safety measures and architectures:

1. classical intra-channel safety mechanisms, such as sensor fusion and RSS [30], can be used as is;
2. if the architecture does not have a multi-channel arbiter or the arbiter's interface is designed differently from ours such as the one proposed in [7], the aggregated safety scores from our cross-channel analysis can still contribute to the Safety Performance Indicators [44] or serve as an independent cross-channel safety monitor output.

# 5. CONCLUSIONS

Redundant multi-channel AD architectures are used in the automotive industry to boost fault tolerance. Our work analyzes the extension of these redundant architectures to also mitigate FIs in the heterogenous AD channels.

To learn about the causes and consequences of FIs, we studied over 2500 DMV disengagement records from the year 2021 and over 10 hours of road test videos of vehicles from 9 different car companies. We observe that FIs of automated driving systems can be just as hazardous as faults, if not more dangerous. Furthermore, FIs are much more frequently the reason of ADS disengagement than system faults based on our observation and data analysis. Our analysis reveals that FIs are problematic and frequently occur not only in the world model generation involving perception and sensor fusion, but also in other AD algorithms with uncertainty, such as motion planning, localization, and vehicle control. To systematically perform SOTIF analysis of triggering conditions and enable development of FI mitigation techniques, we present a novel classification and characterization of OIs based on the available road test data.

Furthermore, we survey state-of-the-art techniques to mitigate FIs and formulate a hypothesis that in an ADS with multiple heterogenous channels, there exist times when some AD channels encounter FIs while the others do not. Our open-loop simulations with three realistic AD channels (Apollo 5.0, Autoware.Auto, and openpilot) support our hypothesis demonstrating that in the same driving scenario different AD channels can be complementary to each other in terms of capabilities to drive safely. Thus, by evaluating the safety of each channel via cross-channel analysis and switching to a sufficiently safe channel at runtime, the ADS safety and availability can be increased. Finally, we propose and discuss a generic architectural design pattern, called Daruma, integrating cross-channel analysis, safety fusion and arbiter functions. The Daruma architectural design pattern reuses state-of-the-art redundant AD architectures typically deployed for fault tolerance to mitigate FIs, while preserving the benefits of the intra-channel FI mitigation techniques in prior works. The Daruma design pattern also allows scaling up safety without sacrificing AD availability, while providing rich opportunities to mitigate many categories of FIs.

## 6. LIMITATIONS AND FUTURE WORK

In this paper we laid the groundwork for a novel conceptual framework for mitigation of the impact of FIs. However, we recognize that our analytical and experimental research had several limitations:

1. Some public road test videos were published by independent users, other videos were published by the company developing the ADS. The latter may have selected more favorable scenarios for the ADS in question, which can skew the FI statistics in our work.
2. The AD channels used in our open-loop simulation experiments have limited features, which may have affected the occurrences of FIs.
3. The triggering condition and the origin of the FIs are hard to pinpoint in our study due to limited available information visible in the public road test and on the ADS displays.

Future research directions include study of FIs related to traffic rules, evaluation of the cross-channel analysis architecture pattern in simulations and road tests, and algorithmic research of the cross-channel analysis and safety fusion. Also, we can update our OI characterization by studying newly released road tests reports, such as the initial data on safety performance of advanced vehicle technologies released by the U.S. National Highway Traffic Safety Administration (NHTSA) [45], and experimenting with more advanced AD channels when they become available.

## ACKNOWLEDGEMENTS

We thank Gerardo Daalderop and Chenchen Dai from NXP Semiconductors for fruitful discussions about the presented work. This work was partially funded by the Dutch Research Council (NWO) through the NEON project.